\documentclass[conference]{IEEEtran}
\IEEEoverridecommandlockouts

\usepackage[utf8]{inputenc}
\usepackage[T1]{fontenc}
\usepackage[english]{babel}
\usepackage{amsmath, amsfonts, amssymb}
\usepackage{algpseudocode}
\usepackage{algorithm}
\usepackage{array}
\usepackage{textcomp}
\usepackage{stfloats}
\usepackage{verbatim}
\usepackage{graphicx}
\usepackage{cite}
\usepackage{import}
\usepackage{bm}
\usepackage{setspace}
\usepackage{dsfont}
\usepackage{indentfirst}
\usepackage[hidelinks]{hyperref}
\usepackage{mathtools}
\usepackage{extarrows}
\usepackage{enumitem}
\usepackage{amsthm}
\usepackage{dsfont}
\usepackage[caption=false]{subfig}
\usepackage[export]{adjustbox}
\usepackage{enumitem}
\usepackage{tcolorbox}
\usepackage[all]{xy}
\usepackage{booktabs}
\usepackage{float}
\usepackage{multirow}
\usepackage[switch]{lineno}

\usepackage{tikz}
\usepackage{pgfplots}
\usetikzlibrary{calc}
\pgfplotsset{compat=newest}

\makeatletter 
\@mparswitchfalse%
\makeatother
\reversemarginpar
\usepackage[%
colorinlistoftodos,prependcaption,textsize=tiny
]{todonotes}

\newcommand{\rbrk}[1]{\left( #1 \right)}
\newcommand{\cbrk}[1]{\left\{ #1 \right\}}
\newcommand{\sbrk}[1]{\left[ #1 \right]}

\newcommand{\trans}{{\mathsf{T}}}

\newcommand{\dagstateset}{\mathcal{S}}
\newcommand{\dagactionset}{\mathcal{A}}
\newcommand{\state}{s}
\newcommand{\terminalset}{\mathcal{X}}
\newcommand{\terminalobject}{x}
\newcommand{\reward}{R}
\newcommand{\marginal}{\pi}
\newcommand{\trajectory}{\tau}
\newcommand{\forwardpolicy}{P_F}
\newcommand{\backwardpolicy}{P_B}
\newcommand{\partitionfunction}{Z}
\newcommand{\gflownetparameters}{\theta}

\newcommand{\dataset}{\mathcal{D}}
\newcommand{\datasetbatch}{\mathcal{B}}
\newcommand{\sample}{x}
\newcommand{\labels}{y}
\newcommand{\utility}{f}
\newcommand{\surrogate}{h}
\newcommand{\gpmean}{m}
\newcommand{\gpkernel}{k}
\newcommand{\gpkernelmatrix}{\mathbf{K}}
\newcommand{\maternscale}{l}
\newcommand{\maternsmoothness}{\nu}

\newcommand{\setdevices}{U}
\newcommand{\deviceindex}{d}
\newcommand{\setmecs}{M}
\newcommand{\mecindex}{m}
\newcommand{\setsubcarriers}{W}
\newcommand{\subcarrierindex}{w}
\newcommand{\bandwidth}{\omega}
\newcommand{\associationmatrix}{\mathbf{X}}
\newcommand{\association}{X}
\newcommand{\bitrate}{\varrho}
\newcommand{\sinr}{\psi}
\newcommand{\transmitpower}{P}
\newcommand{\channel}{g}
\newcommand{\noisepower}{\sigma}
\newcommand{\sensingwaveform}{\alpha}
\newcommand{\reflectedsensingwaveform}{z}
\newcommand{\impulseresponsesensingwaveform}{q}
\newcommand{\fftimpulseresponsesensingwaveform}{Q}
\newcommand{\noisesensingwaveform}{n}
\newcommand{\sensingrate}{\zeta}
\newcommand{\consecutivesensingsymbols}{O}
\newcommand{\sensingsymbolduration}{T_o}
\newcommand{\devicetodevicechannel}{g}
\newcommand{\devicecomputingload}{\lambda}
\newcommand{\meccomputing}{\varepsilon}
\newcommand{\devicedelay}{\xi}
\newcommand{\utilityweights}{\omega}
\newcommand{\nrounds}{N}
\newcommand{\dklparameters}{\phi}

\usepackage{soul, xcolor}

\usepackage[acronym]{glossaries}

\begin{document}


\bstctlcite{IEEEexample:BSTcontrol}

\newacronym{ml}{ML}{machine learning}
\newacronym{ai}{AI}{artificial intelligence}
\newacronym{csi}{CSI}{channel state information}
\newacronym{jepa}{JEPA}{joint-embedding predictive architecture}
\newacronym{jea}{JEA}{joint-embedding architecture}
\newacronym{ofdm}{OFDM}{orthogonal frequency division multiplex}
\newacronym{pdr}{PDR}{pedestrian dead reckoning}
\newacronym{rnn}{RNN}{recurrent neural network}
\newacronym{ema}{EMA}{exponential moving average}
\newacronym{mlp}{MLP}{multi-layer preceptron}
\newacronym{gru}{GRU}{gated recurrent unit}
\newacronym{lstm}{LSTM}{long short-term memory}
\newacronym{1nn}{$1$--NN}{nearest neighbor}
\newacronym{adp}{ADP}{angle-delay profile}
\newacronym{wjepa}{W-JEPA}{}
\newacronym{mdp}{MDP}{Markov decision process}
\newacronym{snr}{SNR}{signal-to-noise ratio}
\newacronym{rssm}{RSSM}{recurrent state-space model}
\newacronym{rl}{RL}{reinforcement learning}
\newacronym{gflownet}{GFlowNet}{generative flow network}
\newacronym{dag}{DAG}{directed acyclic graph}
\newacronym{mec}{MEC}{mobile edge computing}
\newacronym{sinr}{SINR}{signal-to-interference and noise ratio}
\newacronym[\glslongpluralkey={Gaussian processes}]{gp}{GP}{Gaussian process}

\glsdisablehyper

\title{%
GFlowNets for Active Learning Based Resource Allocation in Next Generation Wireless Networks
}

\author{
\IEEEauthorblockN{%
Charbel Bou Chaaya and Mehdi Bennis
}
\IEEEauthorblockA{%
Centre for Wireless Communications, University of Oulu, Finland\\
Emails: \{charbel.bouchaaya, mehdi.bennis\}@oulu.fi}
}

\maketitle

\begin{spacing}{0.95}

\begin{abstract}
In this work, we consider the radio resource allocation problem in a wireless system with various integrated functionalities, such as communication, sensing and computing.
We design suitable resource management techniques that can simultaneously cater to those heterogeneous requirements, and scale appropriately with the high-dimensional and discrete nature of the problem.
We propose a novel active learning framework where resource allocation patterns are drawn sequentially, evaluated in the environment, and then used to iteratively update a surrogate model of the environment.
Our method leverages a \gls{gflownet} to sample favorable solutions, as such models are trained to generate compositional objects proportionally to their training reward, hence providing an appropriate coverage of its modes.
As such, \gls{gflownet} generates diverse and high return resource management designs that update the surrogate model and swiftly discover suitable solutions.
We provide simulation results showing that our method can allocate radio resources achieving 20\% performance gains against benchmarks, while requiring less than half of the number of acquisition rounds.
\end{abstract}

\begin{IEEEkeywords}
Radio resource management, integrated sensing communication and computing, active learning, generative flow network.
\end{IEEEkeywords}

\glsresetall

\section{Introduction}
\IEEEPARstart{T}{he} emergence of novel wireless applications, where communication links are leveraged for sensing and computing, necessitates new radio resource management designs.
In fact, in next generation wireless networks, involving semantic communications and remote control, devices actively sense their surroundings to acquire situational awareness, and query edge servers with sufficient computing capabilities to offload their demanding tasks~\cite{chafii2023twelve}.
Radio resource allocation schemes entailed by such systems fundamentally diverge from classical resource optimization techniques involving zero-sum game trade-offs, where the optimization of a utility degrades another one (for e.g., bitrate and reliability)~\cite{bennis2018ultrareliable}.
In essence, modern applications require the simultaneous satisfaction of high utilities in terms of communication, sensing and computing to cater for various demands, guaranteeing immersive experiences~\cite{wen2024survey}.

Tackling such problems has been previously attempted from multiple directions in the literature.
In~\cite{zhao2022radio}, the resource allocation problem for a network comprising devices with heterogeneous utilities is solved using matching theory, while~\cite{he2023integrated} proposed to decouple the joint optimization of multiple utilities using a threshold policy.
Similar problems involving beamforming designs and resource management have been studied using classical optimization methods in~\cite{li2023integrated} and~\cite{zhuang2023integrated}, whereas \cite{wang2024mean} considered the case where the number of devices scales largely using mean field game theory.
To avoid the complexity of classical techniques, data-driven deep \gls{rl} algorithms have been proposed in~\cite{zhang2021joint} and~\cite{yang2024deep}.
Although such approaches exploit the ability of modern neural networks to learn intricate functions and solve challenging tasks, \gls{rl} methods typically fail to efficiently handle large action spaces.
This is due to their training objective, seeking a policy that maximizes the expected return, which discourages sufficient exploration beyond local optima, particularly in high-dimensional and discrete sets which are common in resource allocation problems.
Recently, graph diffusion models have been used in~\cite{liang2024gdsg} to solve computation offloading problems.
However, such models necessitate lengthy epochs to converge incurring significant training delay, and requiring frequent re-training to adapt to varying wireless environments.

We overcome the drawbacks of previous works by re-casting multi-functional radio resource management as an \emph{active learning} problem~\cite{aggarwal2014active},~\cite{maggi2021bayesian}, where a learning agent sequentially learns to optimize the distribution of resources.
In particular, we use a \gls{gflownet}~\cite{bengio2021flow} to sample diverse high-return allocation patterns through discrete probabilistic modeling.
\Glspl{gflownet} have shown encouraging results when deployed in an active learning loop to quickly discover favorable high-dimensional solutions, as they are trained to generate samples proportionally to their reward, hence very likely sampling from the modes of the reward function.
Their applications have been mostly related to biological drug designs~\cite{jain2022biological},~\cite{hernandezmulti}.
We report several gains of \glspl{gflownet} in radio resource management problems, significantly reducing the number of sampled solutions while achieving $20\%$ performance gains compared to baselines.

\section{Preliminaries -- GFlowNets \& Active Learning}
\subsection{Generative Flow Networks}\label{section:gfnal_conference_gflownets}
\Glspl{gflownet} are a family of probabilistic methods that sample compositional objects from discrete unnormalized distributions~\cite{bengio2023gflownet}.
The sampling problem is treated as a sequential decision making problem, where a compositional object is generated incrementally by a sequence of actions.
Formally, we consider a \gls{dag} $\rbrk{\dagstateset, \dagactionset}$ where $\dagstateset$ is a finite set of states, and $\dagactionset \subseteq \dagstateset \times \dagstateset$ is the set of actions, which represent transitions (directed edges) between states $\state_t \to \state_{t+1}$.
We assume the existence of a unique initial state $\state_0$ with no incoming edges, whereas states with no outgoing edges are terminal states are terminal states forming a set $\terminalset$.
Let $\reward\!\!\!: \terminalset \to \mathbb{R}^+$ be a non-negative reward function defined over terminal states.
The aim of a \gls{gflownet} is to sample objects $\terminalobject \in \terminalset$ with a probability proportional to their reward $\marginal\rbrk{\terminalobject} \propto \reward\rbrk{\terminalobject}$, which can be seen as a unnormalized probability mass.

To approximate this sampling problem, an object $\terminalobject \in \terminalset$ is sequentially constructed by drawing stochastic transitions between partially constructed states starting from $\state_0$, hence forming a complete trajectory $\trajectory = \rbrk{\state_0 \to \state_1 \to \dots \to \state_n}$, where $\forall i,\, \rbrk{\state_i \to \state_{i+1}} \in \dagactionset$ and $\state_n=\terminalobject \in \terminalset$.
As such, a \gls{gflownet} trains a stochastic forward policy $\forwardpolicy\rbrk{\state^\prime \mid \state}$ modeling the distribution over transitions from a non-terminal state.
The forward policy induces a distribution over complete trajectories via $\forwardpolicy\rbrk{\trajectory} = \prod_{i=1}^{n} \forwardpolicy\rbrk{\state_{i} \mid \state_{i-1}}$.
Likewise, the backward policy is a distribution $\backwardpolicy\rbrk{\state \mid \state^\prime}$ over the parents of a non-initial state.
Notice that the probability of generating an object $\terminalobject \in \terminalset$ is $\marginal\rbrk{\terminalobject} = \sum_{\trajectory	\rightsquigarrow\terminalobject} \forwardpolicy\rbrk{\trajectory}$, which is the marginal likelihood of sampling trajectories ending with $\terminalobject$.

The \gls{gflownet} objective is to satisfy $\marginal\rbrk{\terminalobject} = \frac{\reward\rbrk{\terminalobject}}{\partitionfunction}$, where $\partitionfunction=\sum_{\terminalobject \in \terminalset} \reward\rbrk{\terminalobject}$ denotes the partition function.
This objective can be re-stated, for any trajectory $\trajectory \rightsquigarrow \terminalobject$, as:
\begin{equation}\label{eq:gfnal_conference_gfn_objective}
    \partitionfunction \, \prod_{i=1}^{n} \forwardpolicy\rbrk{\state_{i} \mid \state_{i-1}} = \reward\rbrk{\terminalobject} \prod_{i=1}^{n} \backwardpolicy\rbrk{\state_{i-1} \mid \state_{i}}.
\end{equation}
The trajectory balance objective~\cite{malkin2022trajectory} transforms this goal into the following loss function:
\begin{equation}\label{eq:gfnal_conference_gfn_loss}
    \ell\rbrk{\trajectory ; \gflownetparameters} = \sbrk{\log\frac{\partitionfunction_{\gflownetparameters} \, P_{F_\gflownetparameters} \rbrk{\trajectory}}{\reward\rbrk{\terminalobject} P_{B_\gflownetparameters} \rbrk{\trajectory \mid \terminalobject}}}^2,
\end{equation}
where the forward and backward policies are parametrized by trainable parameters $\gflownetparameters$ (for e.g., neural networks), $\partitionfunction_\gflownetparameters$ approximates the partition function, and $\backwardpolicy\rbrk{\trajectory \mid \terminalobject} = \prod_{i=1}^{n} \backwardpolicy\rbrk{\state_{i-1} \mid \state_{i}}$.
When this loss is globally zeroed out for all possible trajectories, \cite[Proposition 1]{malkin2022trajectory} guarantees that $\marginal\rbrk{\terminalobject} \propto \reward\rbrk{\terminalobject}$.
Typically, \gls{gflownet} trajectories are sampled during training, and the objective in~\eqref{eq:gfnal_conference_gfn_loss} is minimized using gradient-based techniques.

\subsection{Active Learning}
Active learning encompasses a set of machine learning methods targeting accelerated training under efficient data sampling schemes~\cite{settles2011theories}.
For instance, unlike classical supervised learning which seeks an accurate model of an unknown function under labeled data, active learning endows the learning agent with the ability to choose which data points to label, hence significantly reducing the need for data annotations.
Such tools are particularly useful when the agent seeks to optimize an unknown function that is expensive to query, as in molecular design problems where experimental lab evaluations of novel drug candidates undergo many screening phases for example.
Hence, instead of labeling large quantities of training data, the agent incrementally learns the target function by actively querying labels for small batches of samples it judges essential.

Typically, we consider an active learning problem where an agent aims to find samples $\sample \in \terminalset$, that maximize a utility $\utility:\terminalset\to\mathbb{R}^+$ that is expensive to evaluate, called the oracle.
Hence, the agent involves two functions: a local surrogate model $\surrogate$ that models the oracle $\utility$ given the observed data, and a sampler $\marginal$ that proposes new samples to evaluate given the proxy $\surrogate$ (for e.g., via an acquisition function).
Practically, active learning rounds proceed as follows: the agent initially observes a dataset $\dataset_0 = \cbrk{\rbrk{\sample_i^0, \labels_i^0}}_{i=1}^n$, where $\labels_i^0=\utility\rbrk{\sample_i^0}$.
At round $j$, the surrogate model is fitted on the data in $\dataset_{j-1}$, and the sampler draws a new batch of sample candidates to be evaluated $\bigl\{\sample_i^j\bigr\}_{i=1}^b$.
The new samples are then annotated by the oracle to form a new batch $\datasetbatch_j = \bigl\{\bigl(\sample_i^j, \labels_i^j\bigr)\bigr\}_{i=1}^b$, which the agent then uses to augment its dataset $\dataset_j = \dataset_{j-1} \cup \datasetbatch_j$, and the procedure is repeated.

A natural choice to model the surrogate function is \glspl{gp}, since they incorporate a notion of uncertainty in their predictions, while retaining a flexible implementation~\cite{rasmussen2005gp}.
Technically, a \gls{gp} is a collection of random variables where the joint distribution of any finite subset is a multivariate Gaussian.
Those random variables are the possible functions modeling the target $\utility$.
Hence, assuming a \gls{gp} prior with mean function $\gpmean$ and covariance kernel $\gpkernel$ over the utility $\utility$, and given a training dataset $\dataset = \cbrk{\rbrk{\sample_i, \labels_i}}_{i=1}^n$ of realizations $\labels_i = \utility\rbrk{\sample_i}$, the utility can be inferred at any point $\sample$ as a Gaussian: $P\rbrk{f\rbrk{\sample} \mid \dataset} \sim \mathcal{N}\rbrk{\mu_\labels, \sigma^2_\labels}$ where:
\begin{align}\label{eq:gfnal_conference_gp_mean_std}
    \mu_\labels &= \gpmean\rbrk{\mathbf{\sample}} + \mathbf{k}_\sample^\trans \, \gpkernelmatrix^{-1} \, \rbrk{\mathbf{\labels} - \gpmean\rbrk{\mathbf{\sample}}},
    \\
    \sigma^2_\labels &= \gpkernel\rbrk{\sample, \sample} - \mathbf{k}_\sample^\trans \, \gpkernelmatrix^{-1} \, \mathbf{k}_\sample,
\end{align}
with $\mathbf{\sample} = \sbrk{\sample_1, \dots, \sample_n}^\trans$, $\mathbf{k}_\sample = \sbrk{\gpkernel\rbrk{\sample, \sample_i}}_i^\trans$ and $\gpkernelmatrix = \sbrk{\gpkernel\rbrk{\sample_i, \sample_j}}_{i,j}$.

\section{System Model and Problem Formulation}
\begin{figure}
    \centering
    \includegraphics[width=.7\linewidth]{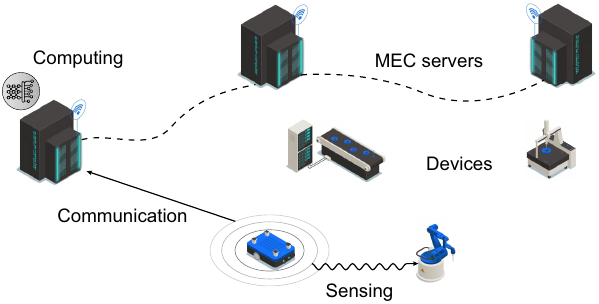}
    \caption{System model.}
    \label{fig:gfnal_conference_system_model}
\end{figure}

%
We consider the uplink of a wireless network, shown in Fig.~\ref{fig:gfnal_conference_system_model}, where a set of $\setdevices$ devices access $\setmecs$ base stations, each of which is a \gls{mec} server.
The communication is managed over $\setsubcarriers$ \gls{ofdm} subcarriers, each spanning a frequency bandwidth $\bandwidth$ used by all base stations.
We assume that each device can be allocated to communicate with one \gls{mec} server over one resource block; conversely, each subcarrier can be retained for one device only.
We denote by $\associationmatrix$ as the $\setmecs \times \setsubcarriers$ resource allocation matrix, whose elements $\association_{\mecindex,\subcarrierindex} = \deviceindex$ indicates that device $\deviceindex$ communicates with \gls{mec} $\mecindex$ over resource block $\subcarrierindex$, and $0$ otherwise.

The devices have diverse tasks ranging from passive data transfer, to situational awareness sensing using integrated signals and running computationally heavy smart applications.
As such, the communication resources must be shared among the users to simultaneously satisfy a favorable overall network performance.
We start by elaborating the performance measure of each device along three continuums: communication, sensing and computing.

\subsection{Communication Model}
We adopt the bitrate as the communication performance metric.
The throughput of device $d$ can be written as:
\begin{equation}\label{eq:gfnal_conference_bitarte}
    \bitrate_\deviceindex = \sum_{\mecindex=1}^\setmecs \sum_{\subcarrierindex=1}^\setsubcarriers \bandwidth \log_2 \rbrk{1 + \sinr_{\deviceindex,\mecindex,\subcarrierindex}},
\end{equation}
where $\sinr_{\deviceindex,\mecindex,\subcarrierindex}$ is the \gls{sinr} of device $\deviceindex$ served by base station $\mecindex$ over subcarrier $\subcarrierindex$.
Given an resource allocation profile $\associationmatrix$, it is expressed as:
\begin{equation}
    \sinr_{\deviceindex,\mecindex,\subcarrierindex} = \frac{\mathds{1}_{\sbrk{\association_{\mecindex,\subcarrierindex} = \deviceindex}} \transmitpower_\deviceindex \, \channel_{\deviceindex,\mecindex,\subcarrierindex}}{\sum_{\deviceindex^\prime \neq \deviceindex} \sum_{\mecindex^\prime=1}^\setmecs \mathds{1}_{[{\association_{\mecindex^\prime,\subcarrierindex} = \deviceindex^\prime}]} \transmitpower_{\deviceindex^\prime} \, \channel_{\deviceindex^\prime,\mecindex,\subcarrierindex} + \noisepower^2},
\end{equation}
where $\transmitpower_\deviceindex$ is the transmit power of device $\deviceindex$, $\channel_{\deviceindex,\mecindex,\subcarrierindex}$ is its uplink channel gain with \gls{mec} $\mecindex$ on subchannel $\subcarrierindex$, and $\mathds{1}_{\sbrk{\cdot}}$ is an indicator function.

\subsection{Sensing Model}
To model its sensing capabilities, we assume that each device transmits integrated \gls{ofdm} waveforms to sense its surrounding environment.
We denote by $\sensingwaveform_{\deviceindex,\subcarrierindex}\rbrk{t}$ as the utilized sensing waveform by device $\deviceindex$ over subcarrier $\subcarrierindex$, which will be reflected by the target for detection.
We assume that the impulse response $\impulseresponsesensingwaveform_{\deviceindex,\subcarrierindex}\rbrk{t}$ of an extended target is a wide-sense stationary Gaussian process.
Therefore, the received signal by the device on subcarrier $\subcarrierindex$ is:
\begin{equation}
    \reflectedsensingwaveform_{\deviceindex,\subcarrierindex}\rbrk{t} = \int_{-\infty}^{\infty} \impulseresponsesensingwaveform_{\deviceindex,\subcarrierindex}\rbrk{\iota} \sensingwaveform_{\deviceindex,\subcarrierindex}\rbrk{t-\iota} \mathrm{d}\iota + \noisesensingwaveform\rbrk{t},
\end{equation}
where the received echos are corrupted by additive white noise $\noisesensingwaveform$ with power $\noisepower^2$.
The sensing performance is then characterized by the conditional mutual information between the target impulse response and the received signal~\cite{liu2017adaptive}:
\begin{align}
    \sensingrate_{\deviceindex,\subcarrierindex} &= \mathrm{I} \rbrk{\reflectedsensingwaveform_{\deviceindex,\subcarrierindex}; \impulseresponsesensingwaveform_{\deviceindex,\subcarrierindex} \mid \sensingwaveform_{\deviceindex,\subcarrierindex}} \\
    &= \frac{1}{2} \, \bandwidth \, \consecutivesensingsymbols \, \sensingsymbolduration \log_2 \rbrk{1 + \sinr^\text{sensing}_{\deviceindex,\subcarrierindex}},
\end{align}
where the sensing \gls{sinr} reads:
\begin{equation}
    \sinr^\text{sensing}_{\deviceindex,\subcarrierindex} = \frac{\sum_{\mecindex=1}^\setmecs \mathds{1}_{\sbrk{\association_{\mecindex,\subcarrierindex} = \deviceindex}} \transmitpower_\deviceindex \, \lvert \fftimpulseresponsesensingwaveform_{\deviceindex,\subcarrierindex} \rvert^2}{\sum_{\deviceindex^\prime \neq \deviceindex} \sum_{\mecindex^\prime \neq \mecindex} \mathds{1}_{[{\association_{\mecindex^\prime,\subcarrierindex} = \deviceindex^\prime}]} \transmitpower_{\deviceindex^\prime} \, \devicetodevicechannel_{\deviceindex,\deviceindex^\prime,\subcarrierindex} + \noisepower^2},
\end{equation}
and we define $\consecutivesensingsymbols$ and $\sensingsymbolduration$ as the number of consecutive \gls{ofdm} symbols and the duration of each symbol respectively, whereas $\devicetodevicechannel_{\deviceindex,\deviceindex^\prime,\subcarrierindex}$ and $\fftimpulseresponsesensingwaveform_{\deviceindex,\subcarrierindex}$ represent the channel gain between two devices and the Fourier transform of $\impulseresponsesensingwaveform_{\deviceindex,\subcarrierindex}$ at the corresponding subcarrier frequency respectively.
Finally, given a resource distribution scheme $\associationmatrix$, a device's sensing performance is: $\sensingrate_\deviceindex = \sum_{\subcarrierindex=1}^\setsubcarriers \sensingrate_{\deviceindex,\subcarrierindex}$.

\subsection{Computing Model}
We denote by $\devicecomputingload_\deviceindex$ as the computational load of device $\deviceindex$, comprising different types of smart applications such as neural network training or remote control operations.
Similarly, $\meccomputing_\mecindex$ represents the computing capacity of \gls{mec} server $\mecindex$.
We let the processing latency experienced by each device serve as its computing performance metric.
It is expressed as:
\begin{equation}\label{eq:gfnal_conference_delay}
    \devicedelay_\deviceindex = \sum_{\mecindex=1}^\setmecs \frac{\sum_{\subcarrierindex=1}^{\setsubcarriers} \mathds{1}_{\sbrk{\association_{\mecindex,\subcarrierindex} = \deviceindex}} \devicecomputingload_\deviceindex \; {\sum_{\subcarrierindex^\prime=1}^\setsubcarriers \sum_{\deviceindex^\prime \neq \deviceindex} \mathds{1}_{[{\association_{\mecindex,\subcarrierindex^\prime} = \deviceindex^\prime}]}}}{\meccomputing_\mecindex}.
\end{equation}
Each edge server's capacity is uniformly split across all the devices' tasks that it serves, given the association matrix $\associationmatrix$.
Note that we ignore the delay incurred by the downstream results re-transmission to the device, as the size of such packets is typically negligible compared to the tasks' inputs (for e.g., label of an input image in classification tasks).
It is also worth mentioning that \eqref{eq:gfnal_conference_delay} neglects the data transmission delay, that is already captured by the device's communication rate in~\eqref{eq:gfnal_conference_bitarte}.
In fact, the data delivery delay is inversely proportional to the device's bitrate.

\subsection{Problem Formulation}
After eliciting our system model, we are now in position to formalize our resource allocation problem.
Given that the devices' requirements are specified along three stems $\rbrk{\bitrate_\deviceindex, \sensingrate_\deviceindex, \devicedelay_\deviceindex}$, we start by defining a scalar utility function, elaborating each end user performance, as follows:
\begin{equation}\label{eq:gfnal_conference_utility}
    \utility_\deviceindex\rbrk{\associationmatrix} = \bitrate_\deviceindex^{\utilityweights^\bitrate_\deviceindex} \times \sensingrate_\deviceindex^{\utilityweights^\sensingrate_\deviceindex} \times \devicedelay_\deviceindex^{-\utilityweights^\devicedelay_\deviceindex},
\end{equation}
where $\utilityweights_\bitrate^\deviceindex, \utilityweights_\sensingrate^\deviceindex$ and $\utilityweights_\devicedelay^\deviceindex$ are positive weights.
The above utility captures the conflicting multi-functional requirements of each device as the product of its communication, sensing and the inverse of its computing metrics.
Further, each term is appropriately weighted by its importance to the device's corresponding task.

As such, we pose our resource optimization problem as follows:
\begin{equation}\label{eq:gfnal_conference_problem}
    \underset{\associationmatrix \in \terminalset}{\text{maximize}} \qquad \utility\rbrk{\associationmatrix} = \sum_{\deviceindex=1}^\setdevices \utility_\deviceindex\rbrk{\associationmatrix},
\end{equation}
where $\terminalset$ is the set of all possible resource allocation schemes.
Problem~\eqref{eq:gfnal_conference_problem} is challenging due to the intricate dependencies between the resource block associations and the user's different utilities on the one hand, and the large search space of possible allocations $\terminalset$ on the other hand.
In fact, problem~\eqref{eq:gfnal_conference_problem} is a variant of the knapsack problem which belongs to the class of NP-complete problems, hence cannot be optimally solved in polynomial time.
Assuming the wireless channels are static during the scheduling interval, and vary independently across different slots, our allocation problem is solved at the beginning of every slot and the distribution of the resources remains fixed during the slot's period.
Thus, it is important to quickly determine high utility solutions at the outset of every slot.

\begin{figure}
    \centering
    \includegraphics[width=.9\linewidth]{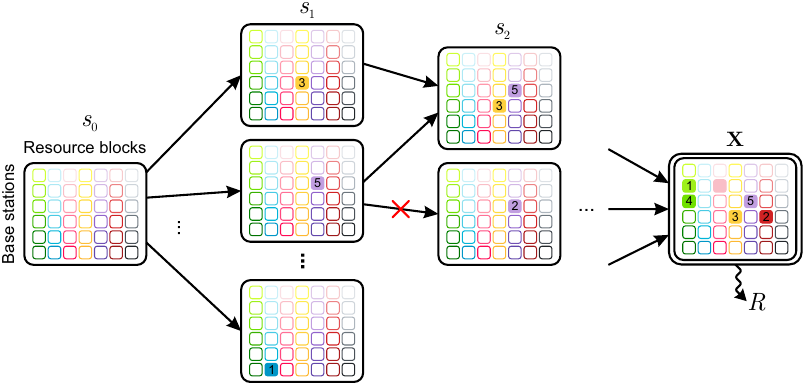}
    \caption{\gls{gflownet} model to sample resource allocation schemes (communication, sensing and computing).}
    \label{fig:gfnal_conference_gflownet_model}
\end{figure}

\section{Generative Active Learning Solution}
%
To solve~\eqref{eq:gfnal_conference_problem}, we propose a data-driven active learning based solution approach, since classical optimization tools fail to appropriately scale their solutions due to the high-dimensional and discrete nature of the problem.
Practically, preparing a dataset where each sample comprises all communication, sensing and computing variables, and labeled by the optimal resource allocation design is unrealistic.
First, this requires exhaustively searching over all possible schemes for every sample scenario, which is computationally expensive, and second, the data used to train a supervised model might not guarantee robust performances after deployment due to distribution shifts.
Hence, to avoid such drawbacks of supervised learning, we develop a framework where the learning agent learns incrementally -- but swiftly -- to discover favorable solutions for~\eqref{eq:gfnal_conference_problem}, therefore quickly configuring the network during the scheduling period.
In particular, since the search space is large, the agent learns to sample high-dimensional solutions from $\terminalset$ with high corresponding utilities $\utility$ by training a generative model, thus we term our approach as generative active learning.

Formally, we treat~\eqref{eq:gfnal_conference_problem} as a black-box optimization problem, where the agent sequentially learns, first fitting a surrogate model $\surrogate$ on a dataset of observations $\cbrk{\sample_i, \labels_i}$ where the samples $\sample_i=\associationmatrix_i$ are labeled by $\labels_i=\utility\rbrk{\sample_i}$, and then sampling new candidates $\sample$ with high expected proxy returns $\surrogate\rbrk{\sample}$ using a generator $\marginal$.
With more sampling rounds, the proxy $\surrogate$ becomes a more accurate model of the expensive utility $\utility$, hence the sampled candidates are more likely to have favorable returns.
We now elaborate our models for the surrogate and the sampler.

\textbf{Surrogate Model.}
We use a \gls{gp} prior for the utility.
To allow for flexible modeling, we use a Matérn class covariance function defined as~\cite{rasmussen2005gp}:
\begin{equation}\label{eq:gfnal_conference_matern}
    \gpkernel\rbrk{\sample_i, \sample_j} = \frac{2^{1 - \maternsmoothness}}{\Gamma\rbrk{\maternsmoothness}} \rbrk{\sqrt{2 \maternsmoothness} \frac{\rvert\sample_i - \sample_j\rvert}{\maternscale}}^\maternsmoothness K_\maternsmoothness \rbrk{\sqrt{2 \maternsmoothness} \frac{\rvert\sample_i - \sample_j\rvert}{\maternscale}},
\end{equation}
where $\Gamma$ is the standard Gamma function and $K_\maternsmoothness$ is a modified Bessel function of the second kind.
Essentially, hyperparameter $\maternsmoothness$ regulates the smoothness of the kernel, while $\maternscale$ is a length-scale variable shaping the function.
In practice, using an exact \gls{gp} does not capture the structure of the high-dimensional space treated by our surrogate.
Hence we rely on deep kernel learning~\cite{wilson2016stochastic}, where a neural network $\dklparameters$ first embeds the samples into low-dimensional representations $\dklparameters\rbrk{\sample_i}$, and the kernel is then applied to the embeddings as $\gpkernel\rbrk{\dklparameters\rbrk{\sample_i}, \dklparameters\rbrk{\sample_j}}$.

\begin{algorithm}[t]
\caption{Generative Active Learning} \label{alg:gfnal_conference_active_learning}
\begin{algorithmic}\setstretch{0.8}
\State\hspace{-0.5em}\textbf{Input}
\State$\utility$: Oracle that evaluates candidates $\sample$ with $\labels$
\State$\surrogate$: Surrogate model of $\utility$
\State$\marginal$: Generative model that samples new candidates $\sample=\associationmatrix$
\State$\dataset_0 = \cbrk{\sample, \labels_i}$: Initial dataset with $\labels_i=\utility\rbrk{\sample}$
\State$\nrounds$: Number of rounds
\State\hspace{-0.5em}\textbf{for} $j=1$ to $\nrounds$ \textbf{do}
\State Fit $\surrogate$ on $\dataset_{j-1}$
\State Train \gls{gflownet} $\marginal$ using reward $\surrogate$
\State Sample batch $\cbrk{\sample}_{i=1}^b$ with $\sample \sim \marginal$
\State Annotate batch with oracle $\utility$: $\datasetbatch_j = \cbrk{\sample, \labels_i}$
\State Update dataset $\dataset_j = \dataset_{j-1} \cup \datasetbatch_j$
\State\hspace{-0.5em}\textbf{end for}
\end{algorithmic}
\end{algorithm}

\begin{figure}
    \centering
    \subfloat[Utility optimization convergence]{
        \includegraphics{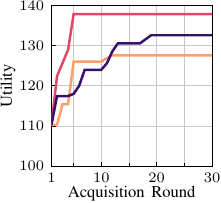}
        \label{fig:gfnal_conference_rounds}}
    \hfill
    \subfloat[Scalability with number of users]{
        \includegraphics{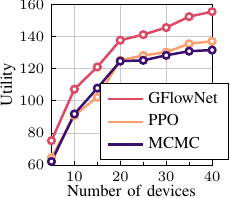}
        \label{fig:gfnal_conference_scalability}}
    \caption{Comparison between different algorithms.}
    \label{fig:gfnal_conference_comparison}
\end{figure}

\begin{figure*}
    \centering
    \subfloat[GFlowNet]{
        \includegraphics{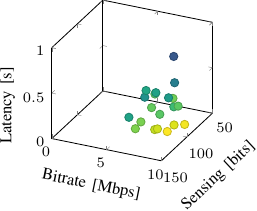}
        \label{fig:gfnal_conference_gfn}}
    \hfill
    \subfloat[PPO]{
        \includegraphics{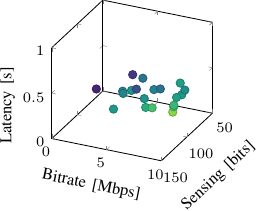}
        \label{fig:gfnal_conference_ppo}}
    \hfill
    \subfloat[MCMC]{
        \includegraphics{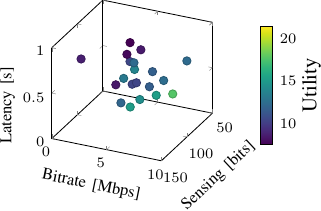}
        \label{fig:gfnal_conference_mcmc}}
    \caption{A sample of solutions found by different methods.}
    \label{fig:gfnal_conference_samples}
\end{figure*}

\textbf{Generator Model.}
We amortize the sampling process of candidates $\associationmatrix$ by training a \gls{gflownet}.
Essentially, we exploit the compositional nature of the resource allocation variable $\associationmatrix$ and train a \gls{gflownet} to sample diverse and high utility solutions from its reward $\reward$, which is updated as the proxy $\surrogate$.
We consider a \gls{dag} whose initial state $\state_0$ is a $\setmecs\times\setsubcarriers$ matrix with zero entries.
At each step, the forward policy modifies one entry of the matrix, setting $\association_{\mecindex,\subcarrierindex}=\deviceindex$ hence connecting device $\deviceindex$ to \gls{mec} $\mecindex$ over resource block $\subcarrierindex$.
Accordingly, the states of the \gls{dag} $\dagstateset$ represent partially distributed allocations, and the actions $\dagactionset$ represent the allocation of a device to a \gls{mec} server on a certain subcarrier.
We constrain the selection process such that previously selected blocks cannot be re-used by other users (given $\mecindex,\subcarrierindex$, $\association_{\mecindex,\subcarrierindex}$ can be selected only once), and the number of entries modified is equal to the number of devices $\setdevices$.
Thus, we guarantee that terminal objects sampled by the \gls{gflownet} are feasible resource allocation matrices $\associationmatrix$.
We illustrate the \gls{gflownet} model in Fig.~\ref{fig:gfnal_conference_gflownet_model}.
We train the model using the trajectory balance loss, reviewed in Section~\ref{section:gfnal_conference_gflownets}.

During its training, a \gls{gflownet} learns to sample $\associationmatrix \sim \marginal$ which is proportional to the reward $\surrogate$, accordingly it learns to generate diverse candidates with favorable returns.
As such, using it as the generator policy to compose the sampled candidates, it provides a good coverage of utility's modes, hence incrementally optimizing~\eqref{eq:gfnal_conference_problem}.
This is in contrast with \gls{rl} tools that rather seek to directly maximize their reward function, and might be stuck in local maxima, particularly when the search space grows largely.
We recapitulate our proposed method in Algorithm~\ref{alg:gfnal_conference_active_learning}.

\section{Simulation Results}
%
To validate our approach, we consider a wireless network with $\setmecs=5$ \gls{mec} servers, each possessing a computational power of $\meccomputing=10$ GHz, communicating over $\setsubcarriers=10$ subcarriers each spanning a bandwidth of $\bandwidth=300$ kHz.
The wireless channel gains are modeled using Rayleigh fading, whereas the devices' task loads are sampled uniformly from $\sbrk{1,2}$ GCPU cycles.
The sensing parameters are set as $\consecutivesensingsymbols=10$ symbols and $\sensingsymbolduration=5 \mu$s.

The \gls{gflownet} policies are parameterized using \glspl{mlp} with two hidden layers of $64$ ReLU activated neurons, and trained with a learning rate of $0.001$.
We compare our candidate sampling method with two baselines: Proximal Policy Optimization (PPO)~\cite{schulman2017proximal}, a deep \gls{rl} technique that is trained to generate candidates maximizing the proxy's return, and Markov chain Monte Carlo (MCMC), a classical heuristic to sample proportionally to an unnormalized density which we implement using the Metropolis-Hastings algorithm.
In each active learning round, $b=16$ samples are proposed by each method, to be evaluated in the environment and update the surrogate model.

In Fig.~\ref{fig:gfnal_conference_rounds}, we show the performance of different algorithms with the number of acquisition rounds, assuming $\setdevices=20$ devices.
We clearly see that the \gls{gflownet} agent finds the best allocation schemes faster than the baselines, due to its ability to generate diverse high-utility candidates.
The PPO method necessitates $10$ rounds to converge, two times more than a \gls{gflownet} whose solution is $8\%$ higher, since the PPO agent halts exploration after $1$ or $2$ modes are found.
The MCMC algorithm continues discovering good candidate solutions, however this requires up to $20$ rounds to find a favorable solution, $4\%$ better than PPO, due to its slow mode mixing time.

In Fig.~\ref{fig:gfnal_conference_scalability}, we vary the number of devices, and report the performance of different methods after $10$ acquisition rounds.
We notice that our \gls{gflownet} technique scales efficiently with the number of users, consistently outperforming benchmarks by around $20\%$.
The performance of PPO and MCMC are more or less similar, both suffering to find favorable solutions in high-dimensional search spaces when the number of devices $\setdevices\geq 20$, with PPO discovering around $5\%$ better allocation schemes.

Finally, Fig.~\ref{fig:gfnal_conference_samples} displays the performance of solutions found by different methods, in terms of per-device communication, sensing and computing metrics.
We observe a much better solution found by the \gls{gflownet} agent, simultaneously guaranteeing overall high metrics in terms of all utilities.

\section{Conclusion}
%
In this work, we proposed an novel active learning approach to allocate radio resources in a wireless system comprising devices with heterogeneous utilities.
We used a \gls{gflownet} to sample high-dimensional allocation patterns ensuring a favorable network performance, proportionally to a reward proxy, which is modeled using a \gls{gp} and successively updated by the generated solutions.
We empirically demonstrated that our method consistently outperforms benchmarks from the literature, quickly discovering high-performing solutions.
Multi-fidelity active learning serves as an extension of our work, where the quality of each candidate annotation is also chosen by the sampler.

\def\baselinestretch{0.82}
\bibliography{references}
\bibliographystyle{IEEEtran}

\end{spacing}

\end{document}